\documentclass{llncs}
\usepackage{named,amsmath}

\def\beq{\begin{equation}}
\def\eeq#1{\label{#1}\end{equation}}
\def\ba{\begin{array}}
\def\ea{\end{array}}

\def\ar{\leftarrow}
\def\rar{\rightarrow}
\def\lrar{\leftrightarrow}
\def\sm{\hbox{\rm SM}}
\def\rv{\hbox{\rm RV}}
\def\ns{\hbox{\rm NS}}
\def\gr{\hbox{\rm Ground}}
\def\raspl{\hbox{RASPL-1}}
\def\i#1{\hbox{\it #1\/}}
\def\no{\i{not}}
\long\def\BOC#1\EOC{\message{(Commented text )}}
\long\def\BOCC#1\EOCC{\message{(Commented text )}}

\newtheorem{prop}{Proposition}
\newtheorem{lem}{Lemma}
\newtheorem{cor}{Corollary}

\title{\bf Safe Formulas\\
in the General Theory of Stable Models}
\author{Joohyung Lee$^1$\inst{} \and
Vladimir Lifschitz$^2$ \and Ravi Palla$^1$\inst{}}
\institute{
$^1$School of Computing and Informatics,
Arizona State University, USA \\
$^2$Department of Computer Sciences,
University of Texas at Austin, USA
{\tt $^1$\{joolee,Ravi.Palla\}@asu.edu,
$^2$vl@cs.utexas.edu}
}

\begin{document}
\date{}
\maketitle

\begin{abstract}
Safe first-order formulas generalize the concept of a safe rule,
which plays an important role in the design of answer set solvers.
We show that any safe sentence is equivalent, in a certain sense, to
the result of its grounding---to the variable-free sentence obtained
from it by replacing all quantifiers with multiple conjunctions and
disjunctions.   It follows that a safe sentence and the result of
its grounding have the same stable models, and that the stable models of
a safe sentence can be characterized by a formula of a simple
syntactic form.
\end{abstract}

\section{Introduction}

The definition of a stable model proposed in \cite{fer07a} is more
general than the original definition from \cite{gel88}: it applies to models of
arbitrary first-order sentences.  Logic programs referred to in the 1988
definition are identified in this theory with first-order formulas of a
special form.  For instance, the rule
\beq
p(x) \ar \no\ q(x)
\eeq{r1}
is treated as alternative notation for the sentence
\beq
\forall x(\neg q(x) \rar p(x)).
\eeq{s1}
In this example, stable models are the interpretations of the unary predicate
constants $p$ and $q$ (in the sense of first-order logic) that make~$p$
identically true and~$q$ identically false.

This general definition of a stable model involves a syntactic transformation
of formulas, which is reviewed in Section~\ref{sec:stable} below.  That
transformation is similar to the circumscription operator \cite{mcc80}---it
turns a first-order sentence into a stronger second-order sentence.  There is
an important difference, however, between stable models and models of
circumscription.
Two sentences may be equivalent (that is, have the same models), but have
different {\it stable} models.  For instance, formula~(\ref{s1}) is equivalent
to
$$
\forall x(\neg p(x) \rar q(x)),
$$
but the stable models of these two formulas are not the same.  The equivalent
transformations of formulas that preserve their stable models are studied
in~\cite{lif07a}.  They are represented there by a  subsystem of classical
logic called ${\bf SQHT}^=$ (``static quantified logic of here-and-there with
equality'').  This deductive system includes all axioms and inference rules of
intuitionistic logic with equality, the decidable equality axiom
\beq
x=y \vee x\neq y
\eeq{de}
and two other axiom schemas, but it does not include the general law of the
excluded middle $F\vee\neg F$.

In \cite{lee08}, the new approach to stable models is used to define
the semantics of an answer set programming language with choice rules and
counting, called \raspl.  The meaning of a {\raspl} program is
defined in terms of the stable models of a first-order sentence associated with
the program, which is called its ``FOL-representation.''  For instance, the
FOL-representation of the {\raspl} rule
\beq
p\ar \{x:q(x)\}\,1
\eeq{r2}
is the formula
\beq
\neg\exists xy(q(x)\land q(y)\land x\neq y) \rar p.
\eeq{s2}

In this paper, we continue one line of research from \cite{lee08}, the
study of safe sentences and their stable models.
%The definition of a safe
%sentence, reproduced in the next section, is related to some ideas of
%\cite{top88}.
It extends the familiar concept of a
safe rule, which plays an important role in the design of answer
set solvers \cite[Section~2.1]{dlv03a}.  For instance, rule~(\ref{r1}) is not
safe, and for this reason it is not allowed in the input of any of the
existing systems for computing stable models.  Rule~(\ref{r2}) is safe, and we
expect that it will be accepted by a future implementation of \raspl.

According to Proposition~\ref{prop-spp} below, stable models of a safe sentence
(without function symbols) have what can be called the ``small predicate
property'': the relation represented by any of its predicate constants can
hold for a tuple of arguments only if each member of the tuple is represented
by an object constant.  We show, furthermore, that any safe sentence is
equivalent, in a
certain sense, to the result of its grounding---to the variable-free sentence
obtained from it by replacing all quantifiers with multiple conjunctions
and disjunctions (Proposition~\ref{prop-qfa}).  We derive from these two facts
that a safe sentence and the result of its grounding have the same stable
models (Proposition~\ref{prop-qfb}).  This theorem leads us to
the conclusion that stable models of a safe sentence can be characterized by
a sentence of a simple syntactic structure---not just first-order, but
universal and, moreover, ``almost variable-free''
(Proposition~\ref{prop-ch}).

The discussion of stable models of safe sentences here is more general than in
\cite{lee08}, because it is not limited to Herbrand models.  This may be
essential for future applications of stable models to knowledge representation.
The theorem about stable Herbrand models stated in \cite{lee08} is now
extended to arbitrary stable models (Proposition~\ref{prop-ext}).

A preliminary report on this work appeared in~\cite{lee08d}.

\section{Review: Stable Models}\label{sec:stable}

The definition of the ``stable model operator'' {\sm} in \cite{fer07a} uses
the following notation from~\cite{lif85}.  Let {\bf p} be
a list of distinct predicate constants $p_1,\dots,p_n$, and let {\bf u} be a
list of distinct predicate variables $u_1,\dots,u_n$ of the same length as
{\bf p}.  By ${\bf u}={\bf p}$ we denote the conjunction of the formulas
$\forall {\bf x}(u_i({\bf x})\lrar p_i({\bf x}))$, where {\bf x} is a list
of distinct object variables of the same arity as the length of $p_i$, for
all $i=1,\dots n$.  By ${\bf u}\leq{\bf p}$ we denote the conjunction of the
formulas $\forall {\bf x}(u_i({\bf x})\rar p_i({\bf x}))$ for all
$i=1,\dots n$, and ${\bf u}<{\bf p}$ stands for
$({\bf u}\leq{\bf p})\land\neg({\bf u}={\bf p})$.
For instance, if~$p$ and~$q$ are unary predicate constants then $(u,v)<(p,q)$
is
$$
\ba l
\forall x(u(x)\rar p(x))\land\forall x(v(x)\rar q(x)) \\
\quad\land\neg(\forall x(u(x)\lrar p(x))\land\forall x(v(x)\lrar q(x))).
\ea
$$

For any first-order sentence $F$, $\sm[F]$
stands for the second-order sentence
\beq
   F \land \neg \exists {\bf u} (({\bf u}<{\bf p}) \land F^*({\bf u})),
\eeq{sm}
where {\bf p} is the list $p_1,\dots,p_n$ of all predicate constants occurring
in $F$, {\bf u} is a list $u_1,\dots,u_n$ of distinct predicate variables,
% of the same length as {\bf p},
and $F^*({\bf u})$ is defined recursively:
\begin{itemize}
\item  $p_i(t_1,\dots,t_m)^* = u_i(t_1,\dots,t_m)$;
\item  $(t_1\!=\!t_2)^* = (t_1\!=\!t_2)$;
\item  $\bot^* = \bot$;
\item  $(F\land G)^* = F^* \land G^*$;
\item  $(F\lor G)^* = F^* \lor G^*$;
\item  $(F\rar G)^* = (F^* \rar G^*)\land (F \rar G)$;
\item  $(\forall xF)^* = \forall xF^*$;
\item  $(\exists xF)^* = \exists xF^*$.
\end{itemize}
An interpretation of the
signature~$\sigma(F)$ consisting of the object and predicate
constants occurring in~$F$ is a {\sl stable model} of~$F$ if it satisfies
$\sm[F]$.

For instance, if $F$ is
\beq
p(a)\land\forall x(p(x)\rar q(x))
\eeq{ex1}
then $F^*(u,v)$ is
$$u(a)\land\forall x((u(x)\rar v(x))\land (p(x)\rar q(x)))$$
and $\sm[F]$ is
$$
\ba l
p(a)\land\forall x(p(x)\rar q(x))\\
\quad\land\neg\exists uv(((u,v)<(p,q))
\land u(a)\land\forall x((u(x)\rar v(x))
% \qquad\qquad\qquad\qquad\qquad\qquad\qquad\qquad
\land(p(x)\rar q(x)))).
\ea
$$
This formula is equivalent to the first-order formula
\beq
\forall x(p(x)\lrar x=a) \wedge \forall x(q(x) \lrar p(x)).
\eeq{ex1comp}
Consequently, the stable models of~(\ref{ex1}) can be characterized as
the interpretations satisfying~(\ref{ex1comp}).

\section{Safe Sentences}\label{sec:safe}

We consider first-order formulas that may contain
object constants and equality but no function constants of arity $>0$.  The
propositional connectives
$$\bot\quad\land\quad\lor\quad\rar$$
will be treated as primitive; $\neg F$ is shorthand for $F\rar\bot$,
$F\lrar G$ is shorthand for $(F\rar G)\land(G\rar F)$, and $\top$ is
shorthand for $\bot\rar\bot$.
A {\sl sentence} is a formula without free variables.

Recall that a traditional rule---an implication of the form
\beq
(L_1\land\cdots\land L_n)\rar A,
\eeq{e1}
not containing equality, where $L_1,\dots,L_n$ are literals and $A$ is an
atom---is considered safe if every variable occurring in it occurs in
one of the positive literals in the antecedent.  The definition of a safe
formula from \cite{lee08}, reproduced below, generalizes this condition to
arbitrary sentences in prenex form.  The assumption
that the formula is in prenex form is not a significant limitation in the
general theory of stable models, because
all steps involved in the standard process of converting a formula to prenex
form are equivalent transformations in ${\bf SQHT}^=$ \cite{lee07}.  For
instance, formula~(\ref{s2}) is equivalent in this system to its prenex form
\beq
\exists xy(\neg(q(x)\land q(y)\land x\neq y) \rar p).
\eeq{s2prenex}

To every quantifier-free formula~$F$ we assign a set $\rv(F)$ of its
{\sl restricted variables} as follows:\footnote{Some clauses of this definition
are similar to parts of the definition of an allowed formula in~\cite{top88}.
That paper was written before the invention of the stable model
semantics, and long before the emergence of answer set programming.}
\begin{itemize}
\item  For an atomic formula~$F$,
  \begin{itemize}
  \item if~$F$ is an equality between two variables then
   $\rv(F)=\emptyset$;
   \item otherwise, $\rv(F)$ is the set of all variables
       occurring in $F$;
  \end{itemize}
\item  $\rv(\bot)=\emptyset$;
\item  $\rv(F\land G)=\rv(F)\cup\rv(G)$;
\item  $\rv(F\lor G)=\rv(F)\cap\rv(G)$;
\item  $\rv(F\rar G)=\emptyset$.
\end{itemize}
We say that a variable $x$ is restricted in~$F$ if $x$ belongs to $\rv(F)$.
It is clear, for instance, that a variable is restricted in the antecedent
of~(\ref{e1}) iff it occurs in one of the positive literals among
$L_1,\dots,L_n$.

Recall that the occurrence of one formula in another is called {\em
positive} if the number of implications containing that occurrence in
the antecedent is even, and {\em negative} otherwise. We say that an
occurrence of a subformula or a variable in a formula~$F$ is
{\em strictly positive} if that occurrence is not in the antecedent of 
any implication. 
For example, in~(\ref{s2}), the occurrences of $q(x)$ and $q(y)$ are
positive, but not strictly positive; the occurrence of $p$ is strictly
positive. 

Consider a sentence~$F$ in prenex form:
\beq
Q_1x_1\cdots Q_nx_nM
\eeq{prenex}
(each $Q_i$ is $\forall$ or $\exists$; $x_1,\dots,x_n$ are distinct
variables; the matrix $M$ is quantifier-free).  
We say that $F$ is {\sl semi-safe} if every strictly positive
occurrence of every variable $x_i$ belongs to a subformula $G\rar H$
where $x_i$ is restricted in $G$.
If a sentence has no strictly positive occurrence of a variable, 
as in~(\ref{s2prenex}), it is clearly semi-safe. For another example,
consider the universal closure of a formula of the
form~(\ref{e1}). If $A$ contains no variables, then the sentence is 
trivially semi-safe. If $A$ contains a variable $x$, then for this sentence
to be semi-safe, $x$ must occur in one of the positive literals among
$L_1,\dots,L_n$.  

Following~\cite{cab09}, we define the following transformations. 
\begin{itemize}
\item  $\neg\bot\ \mapsto\ \top$, \ \ \ $\neg\top\ \mapsto\ \bot$, 
\item  $\bot\land F\ \mapsto\ \bot$, \ \ \ 
       $F\land\bot\ \mapsto\ \bot$, \ \ \ 
       $\top\land F\ \mapsto\  F$, \ \ \
       $F\land\top\ \mapsto\ F$, 
\item  $\bot\lor F\ \mapsto\  F$,\ \ \ 
       $F\lor\bot\ \mapsto\ F$,\ \ \ 
       $\top\lor F\ \mapsto\ \top$,\ \ \ 
       $F\lor\top\ \mapsto\ \top$, 
\item  $\bot\rar F\ \mapsto\ \top$,\ \ \ 
       $F\rar\top\ \mapsto\ \top$, \ \ \ 
       $\top\rar F\ \mapsto\ F$. 
\end{itemize}
Note that these transformations result in a formula that preserves 
equivalence in ${\bf INT}^=$.

We say that a variable $x$ is {\em positively weakly restricted} in a formula
$G$ if the formula obtained from $G$ by 
\begin{itemize}
\item  first replacing every atomic formula $A$ in it such that 
       $x$ is restricted in $A$ by~$\bot$, 
\item  and then applying the transformations above
\end{itemize}
is $\top$. Similarly, we say that $x$ is {\em negatively weakly 
restricted} in $G$ if the formula obtained from $G$ by the same procedure 
is $\bot$. 

We say that a semi-safe sentence~(\ref{prenex}) is {\sl safe} if, for every 
occurrence of a variable~$x_i$ in (\ref{prenex}),
\begin{itemize}
\item[(a)] if $Q_i$ is $\forall$, then the occurrence belongs to 
  \begin{itemize}
  \item a positive subformula of~(\ref{prenex}) in which $x_i$ is
        positively weakly restricted, or
  \item a negative subformula of~(\ref{prenex}) in which $x_i$ is 
        negatively weakly restricted; 
  \end{itemize}
\item[(b)] if $Q_i$ is $\exists$, then the occurrence belongs to 
  \begin{itemize}
  \item a negative subformula of~(\ref{prenex}) in which $x_i$ is
        positively weakly restricted, or
  \item a positive subformula of~(\ref{prenex}) in which $x_i$ is 
        negatively weakly restricted.
  \end{itemize}
\end{itemize}

Consider again the universal closure of a formula of the
form~(\ref{e1}).  If each of its variables occurs in a positive literal in the
antecedent then the matrix~(\ref{e1}) plays the role of the positive
subformula from the definition of a safe sentence.
For another example, sentence~(\ref{s2prenex}) is also safe because the
antecedent of the implication is a negative subformula in which 
both $x$ and $y$ are positively weakly restricted in it. (Or 
$q(x)\land q(y)\land x\neq y$ can be taken as a positive subformula 
in which $x$ and $y$ are negatively weakly restricted.)
Formula
$$\exists x\forall y ((p(x)\rar q(y))\rar r)$$
is safe because, for $x$, $p(x)$ can be taken as a positive subformula, 
and, for $y$, $q(y)$ can be taken as a negative subformula. 
Formula 
$$\exists x(\neg p(x)\rar q)$$
is safe, while 
$$\forall x(\neg p(x)\rar q),$$
is semi-safe, but not safe. 

%[ more examples; comparison with Cabalar et al. ]

\section{The Small Predicate Property}\label{sec:spp}

Proposition~\ref{prop-spp} below shows that all stable models of a safe
sentence have the small predicate property: the relation
represented by any of its predicate constants~$p_i$ can hold for a tuple of
arguments only if each member of the tuple is represented by an object
constant occurring in~$F$.  To make this idea precise, we will use the
following notation: for any finite set~{\bf c} of object constants,
$\i{in}_{\bf c}(x_1,\dots,x_m)$ stands for the formula
$$\bigwedge_{1\leq j\leq m}\;\bigvee_{c\in{\bf c}}\;x_j=c.$$
The small predicate property can be expressed by the conjunction of the
sentences
$$\forall {\bf x} (p_i({\bf x})\rar \i{in}_{\bf c}({\bf x}))$$
for all predicate constants $p_i$ occurring in~$F$,
where~{\bf x} is a list of distinct variables.  We
will denote this sentence by $\i{SPP}_{\bf c}$.
By $c(F)$ we denote the set of all object constants occurring in~$F$.

\begin{prop}\label{prop-spp}
For any semi-safe sentence~$F$, $\sm[F]$ entails $\i{SPP}_{c(F)}$.
\end{prop}

For instance, in application to the prenex form of~(\ref{ex1}) this
proposition asserts that  $\sm[F]$ entails
$$\forall x(p(x)\rar x=a) \wedge \forall x(q(x) \rar x=a).$$

\begin{cor}
For any semi-safe sentence~$F$ that does not contain object constants, $\sm[F]$
entails the formulas $\forall{\bf x}\neg p_i({\bf x})$ for all predicate
constants~$p_i$ of arity $>0$.
\end{cor}

Indeed, $\i{SPP}_\emptyset$ is equivalent to the conjunction of all these
formulas.

\BOC
We will show now how to prove a more general form of
Proposition~\ref{prop-spp}, in which the safety assumption is weakened.
%The definition of a ``semi-safe''
%sentence is obtained from the definition of a safe sentence
%(Section~\ref{sec:safe}) by dropping clause~(a).  That is to say,
We call a sentence (\ref{prenex}) in prenex form {\sl semi-safe} if every
strictly positive\footnote{We say that an
occurrence of a subformula or a variable in a formula~$F$ is
{\em strictly positive} if that occurrence is not in the antecedent of
any implication.}
occurrence of each of the variables~$x_i$ in~$M$ is
contained in a subformula $G\rar H$ such that $x_i$ is restricted
in~$G$.
For instance, the sentence
$$\forall x(\neg p(x)\rar q),$$
representing the rule
$$q\ar \no\ p(x),$$
is not safe, but semi-safe because the occurrence of $x$ is not strictly
positive.
The statement of Proposition~\ref{prop-spp} holds for all semi-safe sentences.
\EOC

We will show now how to prove Proposition~\ref{prop-spp}.
The notation that we use in the proof involves {\sl predicate expressions}
of the form
\beq
\lambda {\bf x} F({\bf x}),
\eeq{exp}
where $F({\bf x})$ is a formula.  If $e$ is (\ref{exp})
% and {\bf t} is a tuple
% of terms of the same length as {\bf x} then $E({\bf t})$ stands for the
% $F({\bf t}$.
% For instance, $\lambda x(x=y)(a)$ stands for $a=y$.
and $G(p)$ is a formula containing a predicate constant $p$ of the same arity
as the length of~{\bf x} then $G(e)$ stands for the result of replacing each
atomic part of the form $p({\bf t})$ in $G(p)$ with $F({\bf t})$, after
renaming the bound variables in $G(p)$ in the usual way, if necessary.
For instance, if $G(p)$ is $p(a)\lor p(b)$ then $G(\lambda y(x=y))$ is
$x=a\lor x=b$.  Substituting a tuple {\bf e} of predicate expressions for a
tuple {\bf p} of predicate constants is defined in a similar way.

For any finite set ${\bf c}$ of object constants, by ${\bf e}_{\bf c}$ we
denote the list of predicate expressions
\[
  \lambda{\bf x}(p_i({\bf x})\land \i{in}_{\bf c}({\bf x}))
\]
for all predicate constants $p_i$.

The following two lemmas can be proved by induction on~$F$.  The first of
them is stated as Lemma~5 in~\cite{ferr09}.

\begin{lem}\label{lem:monotonicity}
For any formula $F$,
\[
(({\bf u} \leq {\bf p})\land F^*({\bf u}))\rar F
\]
is logically valid.
\label{lemmaimplies}
\end{lem}

\begin{lem} \label{lem:rv}
For any quantifier-free formula~$F$ and any finite set~{\bf c} of
object constants containing~$c(F)$,
% \beq
$$
   F^*({\bf e}_{\bf c})\rar\i{in}_{\bf c}(\rv(F))
$$
% \eeq{fsrv}
is logically valid.
\end{lem}

About a variable~$x$ occurring in a quantifier-free formula~$F$ we say that
it is {\sl semi-safe}
in~$F$ if every strictly positive occurrence of~$x$ in~$F$ belongs to a subformula~$G\rar H$
such that~$x$ is restricted in~$G$.  It is clear that a sentence in prenex
form is semi-safe iff all variables in its matrix are semi-safe.  By $\ns(F)$
we will denote the set of the variables of~$F$ that are {\it not} semi-safe.

\begin{lem}\label{lem:unsafe}
For any quantifier-free formula $F$ and any finite set~{\bf c} of object
constants containing~$c(F)$,
\beq
   (F\land \i{in}_{\bf c}(\ns(F)))\rar F^*({\bf e}_{\bf c})
\eeq{unsafe}
is logically valid.
\end{lem}

\proof By induction on~$F$. We only consider the case when $F$ is
$G\rar H$; the other cases are straightforward.
By the induction hypothesis,
\beq
   (H\land \i{in}_{\bf c}(\ns(H)))\rar H^*({\bf e}_{\bf c})
\eeq{ih}
is logically valid.
By Lemma~\ref{lem:monotonicity}, since ${\bf e}_{\bf c}\le {\bf p}$,
\beq
G^*({\bf e}_{\bf c})\rar G
\eeq{mm}
is logically valid. By Lemma~\ref{lem:rv},
\beq
   G^*({\bf e}_{\bf c})\rar\i{in}_{\bf c}(\rv(G))
\eeq{rr}
is logically valid.
Assume the antecedent of~(\ref{unsafe})
% \beq
%   G\rar H\ ,
% \eeq{s-gh}
% \beq
% \i{in}_{\bf c}(\ns({G\rar H}))
% \eeq{s-rar-vgh}
\beq
(G\rar H)\land\i{in}_{\bf c}(\ns({G\rar H})).
\eeq{both}
Assume $G^*({\bf e}_{\bf c})$;  our goal is to derive
$H^*({\bf e}_{\bf c})$.
By~(\ref{mm}),~$G$; by the first conjunctive
term of~(\ref{both}),~$H$.  By~(\ref{rr}),
\beq
   \i{in}_{\bf c}(\rv(G)).
\eeq{s-rar-rvg}
Note that $\ns(H)\subseteq\ns({G\rar H})\cup\rv(G)$.
Consequently, from the second conjunctive term of~(\ref{both})
and~(\ref{s-rar-rvg}),
\beq
   \i{in}_{\bf c}(\ns(H)).
\eeq{s-rar-vh}
From $H$, (\ref{s-rar-vh}) and~(\ref{ih}),
$H^*({\bf e}_{\bf c})$.
\qed

\begin{lem}\label{lem:unsafe2}
For any semi-safe sentence $F$ and any finite set~{\bf c} of object
constants containing~$c(F)$, $F$ entails $F^*({\bf e}_{\bf c})$.
\end{lem}

\proof
Immediate from Lemma~\ref{lem:unsafe}.
\qed

\bigskip\noindent
{\bf Proposition~\ref{prop-spp}, Stronger Form}\hspace{1mm}
For any semi-safe sentence~$F$, $\sm[F]$ entails $\i{SPP}_{c(F)}$.

\proof
Assume $F$ and $\neg \i{SPP}_{c(F)}$; we will derive
\[
   \exists {\bf u}({\bf u}<{\bf p}\land F^*({\bf u})).
\]
To this end, we will prove
% \beq
$$
  ({\bf e}_{c(F)}<{\bf p})\land F^*({\bf e}_{c(F)}).
$$
% \eeq{epfe}
By Lemma~\ref{lem:unsafe2}, it is sufficient to prove the first conjunctive
term, that is,
\begin{align}\label{ep}
& \bigwedge_{p\in {\bf p}} \bigg(\forall {\bf x} \Big(p({\bf x})\land
        \i{in}_{c(F)}(\bf x) \rar \i{p}({\bf x})\Big) \bigg)
 \notag \\
& ~~~~~~ \land \neg\bigwedge_{p\in {\bf p}}
           \forall {\bf x}\bigg(p({\bf x})\rar \Big(p({\bf x})\land
                 \i{in}_{c(F)}(\bf x)\Big)\bigg).
\end{align}
The first conjunctive term of (\ref{ep}) is logically valid, and the
second is equivalent to $\neg \i{SPP}_{c(F)}$.
\qed

\section{Grounding}

The process of grounding replaces quantifiers by multiple conjunctions and
disjunctions.  To make this idea precise, we define, for any sentence~$F$
in prenex form and any nonempty finite set {\bf c} of object constants, the
variable-free formula $\gr_{\bf c}[F]$ as follows.  If~$F$ is quantifier-free
then $\gr_{\bf c}[F]=F$.  Otherwise,
$$\gr_{\bf c}[\forall xF(x)]=\bigwedge_{c\in{\bf c}}\gr_{\bf c}[F(c)],$$
$$\gr_{\bf c}[\exists xF(x)]=\bigvee_{c\in{\bf c}}\gr_{\bf c}[F(c)].$$

As in \cite{lif07a}, by ${\bf INT}^=$ we denote intuitionistic predicate
logic with equality, and DE stands for the decidable equality axiom~(\ref{de}).
The importance of the logical system ${\bf INT}^=+\hbox{\rm DE}$ is determined
by the fact that it is a part of ${\bf SQHT}^=$, so that the provability of a
sentence $F\lrar G$ in this system implies that $\sm[F]$ is equivalent
to $\sm[G]$.

\begin{prop}\label{prop-qfa}
For any safe sentence~$F$ and any nonempty finite set~{\bf c} of object
constants containing~$c(F)$, the equivalence
$$\gr_{\bf c}[F]\lrar F$$
is derivable from $\i{SPP}_{\bf c}$ in ${\bf INT}^=+\hbox{\rm DE}$.
\end{prop}

\begin{lem}\label{lemma-subst}
If any of the sentences $\forall xF(x)$, $\exists xF(x)$ is safe then so is
$F(c)$ for any object constant~$c$.
\end{lem}

\proof Immediate from the fact, easily verified by
induction, that if a variable other than~$x$ is restricted in a formula~$G(x)$
then it is restricted in~$G(c)$ as well.
\qed

\begin{lem}\label{lemma-restr}
If~$x$ is restricted in a quantifier-free formula~$F(x)$, and~{\bf c} is a
nonempty finite set of object constants containing~$c(F)$, then the formula
$$F(x)\rar\i{in}_{\bf c}(x)$$
is derivable from $\i{SPP}_{\bf c}$ in ${\bf INT}^=$.
\end{lem}

\proof Immediate by induction on~$F(x)$. \qed

\begin{lem}\label{lemma-wrestr}
Let $F(x)$ be a quantifier-free formula, and let ${\bf c}$ be a nonempty
finite set of object constants containing~$c(F)$. 
\begin{itemize}
\item[(a)] If~$x$ is positively weakly restricted in~$F(x)$, then 
    \hbox{$\neg in_{\bf c}(x)\rar (F(x)\lrar\top)$} is derivable from 
    $\i{SPP}_{\bf c}$ in ${\bf INT}^=$. 
\item[(b)] If~$x$ is negatively weakly restricted in~$F(x)$, then 
    \hbox{$\neg in_{\bf c}(x)\rar (F(x)\lrar\bot)$} is derivable from 
    $\i{SPP}_{\bf c}$ in ${\bf INT}^=$. 
\end{itemize}
\end{lem}

\proof
(a) By Lemma~\ref{lemma-restr}, for any atomic formula $A$ in which $x$ is 
restricted, $\neg\i{in}_{\bf c}(x)\rar (A\lrar\bot)$ is derivable 
from $\i{SPP}_{\bf c}$ in ${\bf INT}^=$. Assume $\neg\i{in}_{\bf c}(x)$. 
Consequently, $F(x)\lrar F(x)_\bot$ is derivable from $\i{SPP}_{\bf c}$ in 
${\bf INT}^=$, where $F(x)_\bot$ is the formula obtained from $F(x)$ by 
replacing its every atomic formula $A$ in which $x$ is restricted
by $\bot$.
Since $x$ is positively weakly restricted
in $F(x)$, formula \hbox{$F(x)_\bot\lrar\top$} is derivable from  
$\i{SPP}_{\bf c}$ in ${\bf INT}^=$, and consequently, so is $F(x)\lrar\top$.

The proof of (b) is similar. 
\qed

\begin{lem}\label{lemma-prop-qfa}
For any formula~$F(x)$ in prenex form that has no free variables other
than~$x$, and for any nonempty finite set {\bf c} of object constants
containing~$c(F)$,
\begin{enumerate}
\item[(a)]
if the sentence~$\forall xF(x)$ is safe then the equivalence
$$\forall x F(x) \lrar \bigwedge_{c\in{\bf c}} F(c)$$
is derivable from $\i{SPP}_{\bf c}$ in ${\bf INT}^=+\hbox{\rm DE}$;
\item[(b)]
if the sentence~$\exists xF(x)$ is safe then the equivalence
$$\exists x F(x) \lrar \bigvee_{c\in{\bf c}} F(c)$$
is derivable from $\i{SPP}_{\bf c}$ in ${\bf INT}^=+\hbox{\rm DE}$.
\end{enumerate}
\end{lem}

% \noindent {\bf Proof.}
\proof
  (a)~Assume that $\forall xF(x)$ is safe.  In ${\bf INT}^=+\hbox{\rm DE}$, 
this formula can be equivalently written as
$$\forall x((\i{in}_{\bf c}(x)\rar F(x))\land(\neg\i{in}_{\bf c}(x)\rar F(x))),$$
and consequently as
\beq
\bigwedge_{c\in{\bf c}} F(c)\;\land\;\forall x(\neg\i{in}_{\bf c}(x)\rar F(x)).
\eeq{pr1}
Consider the maximal positive subformulas $G(x)$ of~$F(x)$ such that $x$ 
is positively weakly restricted in~$G(x)$. 
By Lemma~\ref{lemma-wrestr} (a), for each of these subformulas, 
the implication 
$$\neg\i{in}_{\bf c}(x)\rar (G(x)\lrar\top)$$
is derivable from $\i{SPP}_{\bf c}$ in ${\bf INT}^=$. 
It follows that, under the assumption $\i{SPP}_{\bf c}$,~(\ref{pr1}) can be
equivalently rewritten as
\beq
\bigwedge_{c\in{\bf c}} F(c)\;\land\;\forall x(\neg\i{in}_{\bf c}(x)\rar S_1),
\eeq{pr2}
where~$S_1$ is the formula obtained from~$F(x)$ by replacing each of these
maximal subformulas $G(x)$ with~$\top$.  
Now consider the maximal negative subformulas $H(x)$ of~$S_1$ such that 
$x$ is negatively weakly restricted in~$H(x)$. 
By Lemma~\ref{lemma-wrestr} (b), for each of these subformulas, 
the implication 
$$\neg\i{in}_{\bf c}(x)\rar (H(x)\lrar\bot)$$
is derivable from $\i{SPP}_{\bf c}$ in ${\bf INT}^=$. 
It follows that, under the assumption $\i{SPP}_{\bf c}$,~(\ref{pr2}) can be
equivalently rewritten as
\beq
\bigwedge_{c\in{\bf c}} F(c)\;\land\;\forall x(\neg\i{in}_{\bf c}(x)\rar S_2),
\eeq{pr3}
where~$S_2$ is the formula obtained from~$S_1$ by replacing each of these
maximal subformulas $H(x)$ with~$\bot$.  

We claim that $x$ does not occur in~$S_2$. Indeed, consider any occurrence 
of~$x$ in~$S_1$. Since $\forall xF(x)$ is safe, in view of the construction 
of~$S_1$, that occurrence is in a negative subformula~$H'(x)$ of~$S_1$, 
which is obtained from a negative subformula~$H(x)$ of~$F(x)$ in which $x$ 
is negatively weakly restricted, by replacing some of its subformulas 
by~$\top$; clearly, $x$ is negatively weakly restricted in~$H'(x)$ as well.
%, but does not belong to any of positive subformula in which $x$ is 
%positively weakly restricted. From the construction of $S_1$, it follows 
%that the occurrence of $x$ in $S_1$ is in a negative subformula $H'(x)$, 
%which is obtained from $H(x)$ by replacing some of its subformulas 
%by~$\top$; clearly, $x$ is negatively weakly restricted in $H'(x)$. 
By the construction of $S_2$, a formula that contains $H'(x)$ is replaced 
by~$\bot$.

%from the construction of 
%$S_1$, it follows that that occurrence is in a negative subformula $H(x)$ 
%of $S_1$ in which $x$ is negatively weakly restricted. From the construction 
%of $S_2$, it follows that $x$ does not occur in $S_2$. 

%since $\forall xF(x)$ is 
%safe, for every occurrence of $x$ in $S_1$, there is an occurrence of $x$
%in $F(x)$ that is in a negative subformula $H(x)$ of $F(x)$ in which 
%$x$ is negatively weakly restricted, but is not in any of 
%positive subformula $G(x)$ of $F(x)$ in which $x$ is positively weakly 
%restricted. From the construction of $S_1$, it is not difficult 
%to check that every occurrence of $x$ in $S_1$ is contained in a negative 
%subformula of $S_1$ in which $x$ is negatively weakly restricted.

\BOC
We claim that $x$ does not occur in $S_2$. Indeed, since $\forall xF(x)$ is 
safe, for every occurrence of $x$ in $S_1$, there is an occurrence of $x$
in $F(x)$ that is in a negative subformula $H(x)$ of $F(x)$ in which 
$x$ is negatively weakly restricted, but is not in any of 
positive subformula $G(x)$ of $F(x)$ in which $x$ is positively weakly 
restricted. From the construction of $S_1$, it is not difficult 
to check that every occurrence of $x$ in $S_1$ is contained in a negative 
subformula of $S_1$ in which $x$ is negatively weakly restricted.
\EOC

It follows that $S_2$ can be obtained from $F(c)$ in the same way as it was
obtained from $F(x)$, that is by replacing some subformulas
that are positive in $F(c)$ with $\top$ and then replacing some
subformulas that are negative in the resulting formula with $\bot$. 
Consequently, $F(c)\rar S_2$ is intuitionistically provable, and so is
$$F(c)\rar\forall x(\neg\i{in}_{\bf c}(x)\rar S_2).$$
It follows that the second conjunctive term of~(\ref{pr3}) can be dropped.

(b)~Assume that $\exists xF(x)$ is safe.  In ${\bf INT}^=+\hbox{\rm DE}$, 
this formula can be equivalently written as
$$\exists x((\i{in}_{\bf c}(x)\land F(x))\lor(\neg\i{in}_{\bf c}(x)\land F(x))),$$
and consequently as
\beq
\bigvee_{c\in{\bf c}} F(c)\;\lor\;\exists x(\neg\i{in}_{\bf c}(x)\land F(x)).
\eeq{pr4}
Consider the maximal negative subformulas $G(x)$ of~$F(x)$ in which 
$x$ is positively weakly restricted.
As before, the implications
$$\neg\i{in}_{\bf c}(x)\rar(G(x)\lrar\top)$$
are derivable from $\i{SPP}_{\bf c}$ in ${\bf INT}^=$.
Consequently, under the assumption $\i{SPP}_{\bf c}$,~(\ref{pr4}) can be
equivalently rewritten as
\beq
\bigvee_{c\in{\bf c}} F(c)\;\lor\;\exists x(\neg\i{in}_{\bf c}(x)\land S_1),
\eeq{pr5}
where~$S_1$ is the formula obtained from~$F(x)$ by replacing each of
these maximal subformulas $G(x)$ with~$\top$.  
Now consider the maximal positive subformulas $H(x)$ of~$S_1$ in which 
$x$ is negatively weakly restricted.
As before, for each of these subformulas, 
the implication 
$$\neg\i{in}_{\bf c}(x)\rar (H(x)\lrar\bot)$$
is derivable from $\i{SPP}_{\bf c}$ in ${\bf INT}^=$. 
Consequently, under the assumption $\i{SPP}_{\bf c}$,~(\ref{pr5}) can be
equivalently rewritten as
\beq
\bigvee_{c\in{\bf c}} F(c)\;\lor\;\exists x(\neg\i{in}_{\bf c}(x)\land S_2),
\eeq{pr6}
where~$S_2$ is the formula obtained from~$S_1$ by replacing each of these
maximal subformulas $H(x)$ with~$\bot$.  
Similar to~(a), $x$ does not occur in $S_2$ and it follows that~$S_2$ 
can be obtained from~$F(c)$ in the same way as it
was obtained from~$F(x)$, that is, by replacing some subformulas that are
negative in~$F(c)$ with~$\top$, and then replacing some subformulas
that are positive in the resulting formula with~$\bot$.  
Consequently, the formula $S_2\rar F(c)$ is
intuitionistically provable, and so is
$$\exists x(\neg\i{in}_{\bf c}(x)\land S_2)\rar F(c).$$
It follows that the second disjunctive term of~(\ref{pr5}) can be dropped.
\qed

\medskip\noindent {\it Proof of Proposition~\ref{prop-qfa}.}
By induction on the length of the prefix.  The base case is trivial.  Assume
that $QxF(x)$ is safe.  Case~1: $Q$ is $\forall$.  In view of
Lemma~\ref{lemma-subst}, from the induction hypothesis we can conclude that
$$\gr_{\bf c}[F(c)]\lrar F(c)$$
is derivable from $\i{SPP}_{\bf c}$ in ${\bf INT}^=+\hbox{\rm DE}$ for every
$c\in{\bf c}$.  Consequently
$$\bigwedge_{c\in{\bf c}}\gr_{\bf c}[F(c)]\lrar\bigwedge_{c\in{\bf c}}F(c)$$
is derivable from $\i{SPP}_{\bf c}$ as well.  By the definition of
$\gr_{\bf c}$, the left-hand side is $\gr_{\bf c}[\forall xF(x)]$.  By
Lemma~\ref{lemma-prop-qfa}(a), under the assumption $\i{SPP}_{\bf c}$ the
right-hand side is equivalent in ${\bf INT}^=+\hbox{\rm DE}$
to~$\forall xF(x)$.  Case~2: $Q$ is $\exists$.  Similar, using
Lemma~\ref{lemma-prop-qfa}(b).
\qed

\medskip
It is interesting that without the decidable equality axiom DE, the statement
of Proposition~\ref{prop-qfa} would be incorrect.  The formula
$$\forall x(((x=a\lor x=b) \rar p(x))\vee((x=a\lor x=b) \rar q(x)))$$
can serve as a counterexample.  Indeed, call this formula~$F$, and assume that
\beq
\gr_{\{a,b\}}[F]\lrar F
\eeq{int1}
can be derived from
\beq
\i{SPP}_{\{a,b\}}
\eeq{int2}
in ${\bf INT}^=$.  In this derivation, substitute $\lambda x(x=a)$ for $p$,
and $\lambda x(x=b)$ for $q$.  After this substitution, the right-hand side
of (\ref{int1}) becomes
\beq
\forall x(((x=a\lor x=b) \rar x=a)\vee((x=a\lor x=b) \rar x=b)),
\eeq{int3}
the left-hand side becomes
\beq
\ba c
          (((a=a\lor a=b) \rar a=a)\vee((a=a\lor a=b) \rar a=b))\\
          \land(((b=a\lor b=b) \rar b=a)\vee((b=a\lor b=b) \rar b=b)),
\ea \eeq{int4} and (\ref{int2}) becomes \beq \forall
x((x=a\rar(x=a\lor x=b))\land(x=b\rar(x=a\vee x=b))). \eeq{int5}
Since (\ref{int4}) and (\ref{int5}) can be proved in ${\bf INT}^=$,
it follows that (\ref{int3}) is provable in this system also.
According to the disjunction property of ${\bf INT}^=$, if a
disjunction is provable in ${\bf INT}^=$ then at least one of its
disjunctive terms is provable. Consequently, at least one of the
formulas
$$(x=a\lor x=b) \rar x=a,\ (x=a\lor x=b) \rar x=b$$
is provable in ${\bf INT}^=$.  But this is impossible, because these formulas
are not even logically valid.

Unlike Proposition~\ref{prop-spp}, Proposition~\ref{prop-qfa} will not hold
if we replace ``safe'' in its statement with ``semi-safe.''  For instance,
take~$F$ to be $\forall x\neg\neg p(x,a)$.  The equivalence
$$\neg\neg p(a,a) \;\lrar\;\forall x\neg\neg p(x,a)$$
is not entailed by the small predicate property
$$\forall xy(p(x,y) \rar (x=a \land y=a))$$
even classically.  (Consider an interpretation with a non-singleton universe
in which~$p(x,y)$ is defined as $x=a \land y=a$.)

%On the other hand, the assertion of Proposition~\ref{prop-qfa} will remain
%true if we replace ``safe'' in its statement with ``semi-safe'' {\it and at the
%same time} add the assumption that {\bf c} is a {\it proper} superset
%of~$c(F)$.  The same can be said about the statement of
%Proposition~\ref{prop-qfb} below.

\begin{prop}\label{prop-qfb}
For any safe sentence~$F$ and any nonempty finite set~{\bf c} of object
constants containing~$c(F)$, $\sm[\gr_{\bf c}[F]]$ is equivalent to~$\sm[F]$.
\end{prop}

In the proof we use the following terminology, which generalizes the
concept of a negative literal.  A formula~$F$ is {\sl negative} if every
occurrence of every predicate constant in~$F$ belongs to the antecedent of
an implication.  For any sentence~$F$ and any negative sentence~$G$,
$\sm[F\land G]$ is equivalent to $\sm[F]\land G$ \cite[Theorem~3]{ferr09}.

\medskip\noindent {\it Proof of Proposition~\ref{prop-qfb}.}
By Proposition~\ref{prop-qfa} proved above, the equivalence
$$\gr_{\bf c}[F]\land \i{SPP}_{\bf c} \lrar F\land \i{SPP}_{\bf c}$$
is provable in ${\bf INT}^=+\hbox{\rm DE}$.  Consequently
\begin{center}
$\sm[\gr_{\bf c}[F]\land \i{SPP}_{\bf c}]\;$ is equivalent to $\;\sm[F\land
\i{SPP}_{\bf c}]$.
\end{center}
Since $\i{SPP}_{\bf c}$ is negative, it follows that
\begin{center}
$\sm[\gr_{\bf c}[F]]\land \i{SPP}_{\bf c}\;$ is equivalent to
$\;\sm[F]\land \i{SPP}_{\bf c}$.
\end{center}
In view of Proposition~\ref{prop-spp}
and the fact that $c(F)\subseteq{\bf c}$, the conjunctive term $\sm[F]$ in
the second conjunction entails its other conjunctive term $\i{SPP}_{\bf c}$,
and the latter can be dropped.  Furthermore, $\gr_{\bf c}[F]$ is variable-free
and consequently safe.  It follows by similar reasoning that in the first
conjunction the term $\i{SPP}_{\bf c}$ can be dropped also.
\qed

\section{Characterizing Stable Models of a Safe Sentence}

\begin{prop}\label{prop-ch}
For every safe sentence~$F$ there exists a variable-free formula~$G$
such that~$\sm[F]$ is equivalent to $G\land\i{SPP}_{c(F)}$.
\end{prop}

% \noindent{\bf Proof.}$\;$
\proof
 In view of
Proposition~\ref{prop-spp}, we need to find a  variable-free formula~$G$ such
that $\i{SPP}_{c(F)}$ entails~$\sm[F]\lrar G$.

Case 1: $c(F)=\emptyset$.  Under the assumption $\i{SPP}_\emptyset$, every
atomic part of $\sm[F]$ that contains a predicate constant or variable of
arity $>0$ can be equivalently replaced by~$\bot$.  The result is a
second-order propositional formula, so that it is equivalent to a
propositional formula.

Case 2: $c(F)\neq\emptyset$ and~$F$ is variable-free.  The only
quantifiers in~(\ref{sm}) are the second-order quantifiers
$\exists {\bf u}$.  Clearly
$\i{SPP}_{c(F)}$ entails
$$u_i\leq p_i \;\rar\;
   u_i\leq\lambda {\bf x}\left(\bigvee_{\bf c}\;{\bf x}={\bf c}\right)$$
where {\bf c} ranges over the tuples of members of $c(F)$ of the same length
as~{\bf x}.  Consequently it entails also
$${\bf u}<{\bf p} \;\rar\;
   u_i\leq\lambda {\bf x}\left(\bigvee_{\bf c}\;{\bf x}={\bf c}\right)$$
and
$${\bf u}<{\bf p} \;\rar\; \bigvee_C \left( u_i=\lambda {\bf x}
                            \bigvee_{{\bf c}\in C}{\bf x}={\bf c}\right),$$
where~$C$ ranges over all sets of such tuples.  It follows that under the
assumption $\i{SPP}_{c(F)}$ the quantifiers $\exists {\bf u}$ can be
equivalently replaced by finite disjunctions, with expressions of the form
$\lambda {\bf x}\bigvee_{{\bf c}\in C}{\bf x}={\bf c}$
substituted for the variables~$u_i$.  The result is a variable-free formula
with the required properties.

Case 3: $c(F)\neq\emptyset$ and~$F$ is not variable-free.  The part of
Proposition~\ref{prop-ch} corresponding to Case~2 can be applied to
$\gr_{c(F)}[F]$.  Since the
formulas~$F$ and $\gr_{c(F)}[F]$ contain the same object constants, we can
assert that, for some variable-free formula~$G$, $\i{SPP}_{c(F)}$
entails
$$\sm[\gr_{c(F)}[F]]\lrar G.$$
It remains to observe that, by Proposition~\ref{prop-qfb}, the left-hand side
is equivalent to $\sm[F]$.
\qed

%\medskip
%The assertion of Proposition~\ref{prop-ch} will remain true
%if we replace ``safe'' in its statement with ``semi-safe'' and replace
%$\i{SPP}_{c(F)}$ with $\i{SPP}_{\bf c}$ for an arbitrary
%proper superset ${\bf c}$ of $c(F)$.

\section{Extending a Stable Model}

Let~$I$ be an interpretation of a set of object and predicate
constants, and let~$X$ be a superset of the universe of~$I$.  By the
{\sl extension of~$I$ to~$X$} we mean the interpretation of the same
constants with the universe~$X$ such that each object constant
represents the same object under both interpretations, and each
predicate constant represents the same set of tuples.

\begin{prop}\label{prop-ext}
For any safe sentence~$F$, any interpretation~$I$ of the object and predicate
constants from~$F$, and any superset~$X$ of the universe of~$I$, the extension
of~$I$ to~$X$ is a stable model of~$F$ iff~$I$ is a stable model of~$F$.
\end{prop}

% \noindent{\bf Proof.}$\;$
\proof
 Consider a variable-free formula~$G$ such that
$\sm[F]$ is equivalent to $G\land\i{SPP}_{c(F)}$ (Proposition~\ref{prop-ch}).
It is clear that~$I$ satisfies~$G$ iff the extension
of~$I$ to~$X$ satisfies~$G$, and that~$I$ satisfies~$\i{SPP}_{c(F)}$ iff the
extension of~$I$ to~$X$ satisfies~$\i{SPP}_{c(F)}$.
\qed

\medskip
In the special case when~$I$ is an Herbrand interpretation, this theorem turns
into Proposition 1 from \cite{lee08}.

%\nb{
%Proposition~\ref{prop-ext} can be extended to a semi-safe sentence as well
%but requiring that the universe of an interpretation contain at least one
%``unnamed'' elements: the statement will remain true if $F$ is assumed
%to be a semi-safe sentence, and $I$ is assumed to be an interpretation
%of the object and predicate constants from~$F$ such that the universe
%contains at least one element $\xi$ where $\xi\ne c^I$ for all
%object constants $c$ from $F$.
%}

\section{Relation to Safety by Cabalar, Pearce, Valverde}

\section{Conclusion}

The approach to stable models developed in \cite{fer07a} is richer than
the traditional view not only syntactically, but also
semantically: stable models became now models in the sense of classical logic,
not merely sets of ground atoms.  But the only models referred to in the
definition of {\raspl} are Herbrand models---sets of ground atoms.  That
definition exploits the syntactic generality of the new theory of stable
models, but not its semantic generality.

We expect, however, that future work on applications of stable
models to knowledge representation will demonstrate the usefulness
of non-Herbrand stable models.  Such models allow us to talk about
elements of the universe that are ``unnamed,'' that is, not
represented by ground terms.  They also allow us to
talk about  elements of the universe that may have ``multiple
names'' in the language.  These additional possibilities may be
certainly useful.

In this paper we investigated properties of stable models of safe formulas
in a semantically general situation, not limited to Herbrand models, and
established a few positive results.  We saw, in particular, that grounding
a safe sentence preserves its stable models even in this general case.  We
hope that these theorems will help us in future work on non-Herbrand
answer set programming.

\section*{Acknowledgements}

We are grateful to Paolo Ferraris and anonymous referees for ICLP 2008
for their useful comments on an earlier version of this paper. The
first and the third author were partially supported by the National
Science Foundation under Grant IIS-0839821. The second author was
partially supported by the National Science Foundation under Grant
IIS-0712113.

\bibliographystyle{named}
\bibliography{bib,bib2}
\end{document}